\documentclass[10pt,twocolumn,letterpaper]{article}

\usepackage{wacv}
\usepackage{times}
\usepackage{epsfig}
\usepackage{graphicx}
\usepackage{amsmath}
\usepackage{amssymb}

\usepackage{multirow}
\usepackage{comment}
\usepackage{xspace}
\usepackage{graphicx}
\usepackage{subcaption}
\usepackage{caption}
\usepackage{algorithm}
\usepackage[noend]{algpseudocode}
\usepackage[accsupp]{axessibility}

\usepackage{enumitem}

\newcommand{\bx}{\boldsymbol{x}}

\newcommand{\bz}{\boldsymbol{z}}

\newcommand{\sC}{\mathcal{C}}

\newcommand{\sS}{\mathcal{S}}
\newcommand{\sT}{\mathcal{T}}

\newcommand{\sL}{\mathcal{L}}
\newcommand{\bbO}{\mathbb{O}}

\DeclareMathOperator*{\argmin}{arg\,min}

\newcommand{\reals}{\mathbb{R}}

\makeatletter
\DeclareRobustCommand\onedot{\futurelet\@let@token\@onedot}
\def\@onedot{\ifx\@let@token.\else.\null\fi\xspace}

\def\ie{\emph{i.e}\onedot}

\usepackage{accents}

\usepackage{xcolor}
\usepackage{multirow}

\makeatother
\def\wacvPaperID{427} 

\wacvfinalcopy 

\ifwacvfinal
\fi

\ifwacvfinal
\usepackage[breaklinks=true,bookmarks=false]{hyperref}
\else
\usepackage[pagebackref=true,breaklinks=true,colorlinks,bookmarks=false]{hyperref}
\fi

\begin{document}

\title{Distance-based Hyperspherical Classification for \\Multi-source Open-Set Domain Adaptation}

\author{Silvia Bucci{\thanks{The authors equally contributed to this work.}} \quad
Francesco Cappio Borlino\footnotemark[1] \quad
Barbara Caputo \quad
Tatiana Tommasi \and
Politecnico di Torino, Italy \and
Italian Institute of Technology \and 
\tt\small \{silvia.bucci, francesco.cappio, barbara.caputo, tatiana.tommasi\}@polito.it
}

\maketitle
\thispagestyle{empty}
\newcommand{\OUR}{HyMOS\xspace}

\begin{abstract}

Vision systems trained in closed-world scenarios fail when presented with new environmental conditions, new data distributions, and novel classes at deployment time. How to move towards open-world learning is a long-standing research question. The existing solutions mainly focus on specific aspects of the problem (single domain Open-Set, multi-domain Closed-Set), or propose complex strategies which combine several losses and manually tuned hyperparameters. In this work, we tackle multi-source Open-Set domain adaptation by introducing \OUR: a straightforward model that exploits the power of contrastive learning and the properties of its hyperspherical feature space to correctly predict known labels on the target, while rejecting samples belonging to any unknown class. 
\OUR includes style transfer among the instance transformations of contrastive learning to  
get domain invariance while avoiding the risk of negative-transfer. 
A self-paced threshold is defined on the basis of the observed data distribution 
and updates online during training, allowing to handle the known-unknown separation.
We validate our method over three challenging datasets.
The obtained results show that \OUR outperforms several competitors, defining the new state-of-the-art. Our code is available at \url{https://github.com/silvia1993/HyMOS}.

\end{abstract}

\section{Introduction}

\begin{figure}[tb]
\centering
\includegraphics[width=0.5\textwidth]{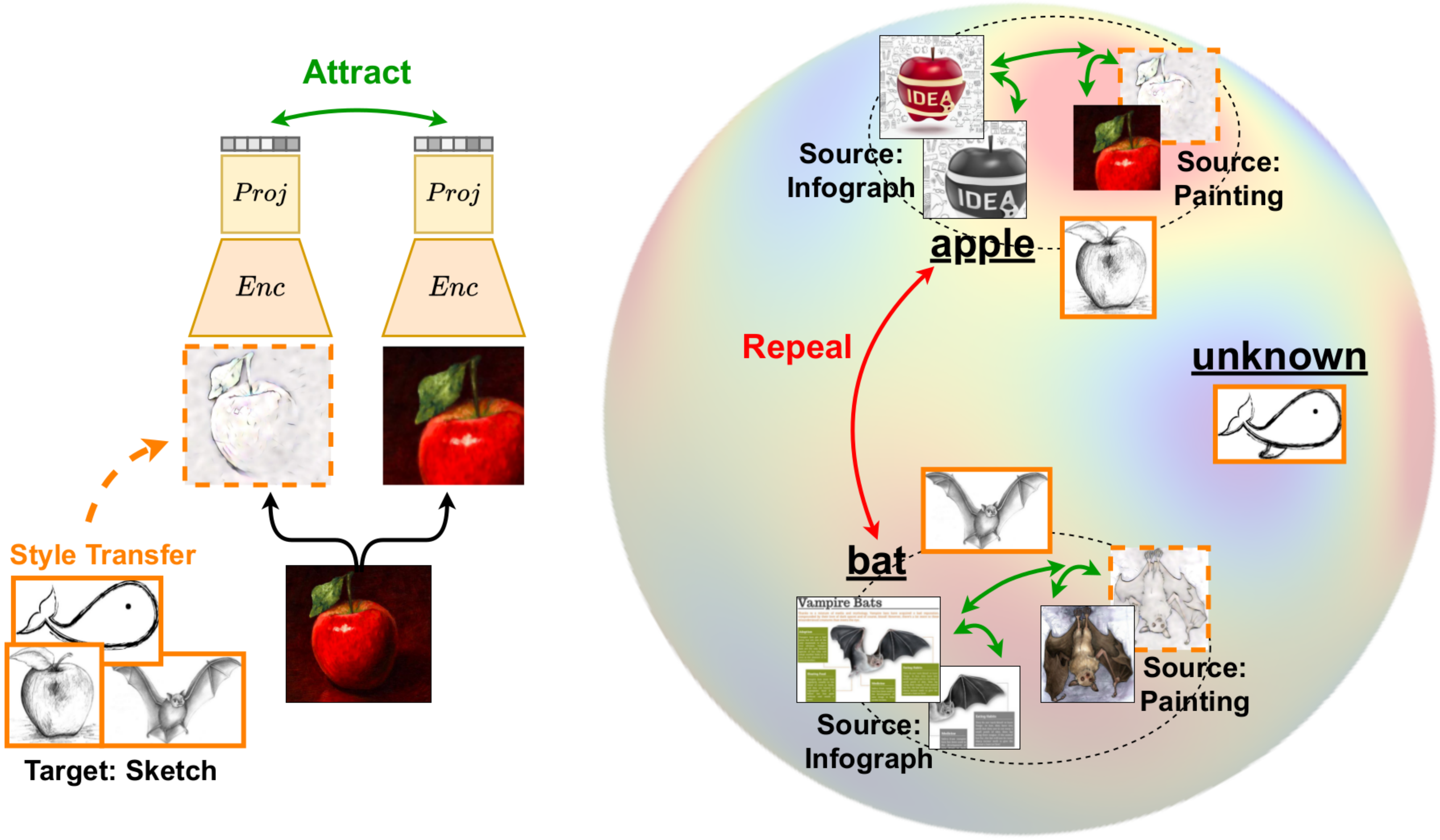}
    \caption{With \OUR we exploit supervised contrastive learning to tackle all the challenges of multi-source Open-Set domain adaptation. We introduce style transfer in the double path contrastive logic to obtain a domain invariant representation. By balancing class and source domains in each training batch we obtain class-wise domain alignment. The learned embedding space naturally isolates target \textit{unknown} samples in low-density regions, while the \textit{known} samples lay close to the corresponding class cluster and can be easily involved in self-training for further adaptation.} \vspace{-3mm}
    \label{fig:new_teaser}
\end{figure}

Artificial intelligent systems face many operational challenges when moving from the controlled lab environment to the real-world.
First of all the annotated data available to train a model might be the result of asynchronous multi-agent collection processes. For vision tasks, this means dealing with datasets composed of labeled images that share the same class set, but with sub-groups of instances showing significant differences in appearance and style among each other. Moreover, at deployment time the learned model will inevitably encounter new environmental conditions, with distribution shift and novel classes not present during training.
In standard Closed-Set domain adaptation \cite{csurka}, the main focus is on how to reduce the gap between the training (labeled source) and test (unlabeled target) data when the latter covers exactly the same class set of the former. 
Open-Set domain adaptation \cite{Busto_2017_ICCV} aims at bridging the domain gap while also rejecting target samples of \textit{unknown} classes. Indeed, in the case of category shift, the application of na\"ive adaptive solutions may lead to \textit{negative-transfer} and unrecoverable class misalignment \cite{liu2019separate}.
Although dealing with multiple sources is more the rule than an exception in real-world conditions, only one recent work has started to peek into the multi-source Open-Set domain adaptation task \cite{rakshit2020multi}. This highlights the difficulty of learning a feature space shared among domains, while also maximizing the separation between \textit{known} and \textit{unknown} categories within the unlabeled target. 

The foundational problem that all the current open-world adaptive learning models try to solve is the limited generalization ability of the albeit powerful deep learning models. This can be at least partially explained considering two well known CNN shortcomings: (1) deep models yield features that describe mostly local rather than global statistics, which causes a bias on the image style of the training data \cite{Jenni_steering}; (2) the cross-entropy loss, widely used for supervised learning, produces overconfident predictions thus biasing the model towards the labeled class set \cite{NEURIPS2020_supclr,lee2017training}.
Existing solutions adopt multi-stage learning procedures,  combine several losses to compensate for the cross-entropy over-reliance on source supervision, and close the domain gap with adversarial techniques.
The obtained approaches are difficult to train with several hyperparameters and manually set thresholds, or include complex models to generate samples of a synthetic \emph{unknown} source class (see Table \ref{tab:comparison}). 


With our work, we propose a supervised model that avoids the drawbacks of the cross-entropy loss, while learning a style-invariant embedding space that naturally isolates the \textit{unknown} categories. Specifically, we build over the very recent contrastive learning trend \cite{simclr2020,MOCO_He_2020_CVPR,NEURIPS2020_supclr}, where the encoder learns the invariance between two augmented views of one image  (positive  pair) while  maximizing the distance among augmented versions of different instances (negative pair). We show how \textbf{a single supervised contrastive learning objective can tackle every challenge of multi-source Open-Set domain adaptation} (see Figure  \ref{fig:new_teaser}):
\begin{itemize}[leftmargin=*]
\vspace{-2mm}\item source to source class-wise alignment comes by simply balancing data batches over classes and domains;
\vspace{-2mm}\item source to target adaptation is obtained by first getting domain invariant features via the introduction of style transfer among the augmentations of contrastive learning. Then, a progressive and auto-regulated self-training procedure further improves the alignment between the target and the source classes clusters;
\vspace{-2mm}\item the separation between \textit{known} and \textit{unknown} target data comes from a self-paced threshold based on the observed data distribution on the learned hyperspherical feature embedding. Indeed, the contrastive objective leads to compact and well separated \textit{known} class clusters \cite{icml_align_uniform}, leaving \textit{unknown} samples isolated in low-density regions.
\end{itemize}

To highlight the important role of the \textbf{Hy}perspherical feature space for our \textbf{M}ulti-source \textbf{O}pen-\textbf{S}et approach, we dub it \textbf{HyMOS}.
We present an extensive experimental analysis on three multi-source Open-Set datasets, showing how \OUR outperforms current state-of-the-art methods. A thorough ablation study provides details on the internal functioning of the method. Further application to related challenging scenarios (multi-source closed set and multi-source universal) show promising results.

\begin{table}[tb]
\begin{center}
\resizebox{\linewidth}{!}{
\begin{tabular}{|c@{~}c@{~~~}c@{~~~}c@{~~~}|}
\hline
\multirow{2}{*}{Method}
& \multicolumn{1}{|@{~}c@{~}|}{No. of} &
\multicolumn{1}{|@{~}c@{~}}{No. of} & \multicolumn{1}{|@{~}c@{~}|}{\multirow{2}{*}{Threshold}} \\

& \multicolumn{1}{|@{~}c@{~}|}{Losses } &
\multicolumn{1}{|@{~}c@{~}}{HPs} & \multicolumn{1}{|@{~}c@{~}|}{} \\

\hline
\multicolumn{1}{|@{~}c@{~}|}{Inheritable \cite{kundu2020towards}} &  \multicolumn{1}{@{~}c@{~}|}{4} &	\multicolumn{1}{@{~}c@{~}|}{2}&	\multicolumn{1}{@{~}c@{~}|}{ not used - synthesize \textit{unknown} target}
\\
\multicolumn{1}{|@{~}c@{~}|}{ROS \cite{bucci2020effectiveness}} & 	\multicolumn{1}{@{~}c@{~}|}{6} & 	\multicolumn{1}{@{~}c@{~}|}{4} & 	\multicolumn{1}{@{~}c@{~}|}{reject a fixed portion of Target} 
\\
\multicolumn{1}{|@{~}c@{~}|}{CMU \cite{fu2020learning}} & 	\multicolumn{1}{@{~}c@{~}|}{$2+|\sC_s|$} & 	\multicolumn{1}{@{~}c@{~}|}{3} & 	\multicolumn{1}{@{~}c@{~}|}{validated} 
\\
\multicolumn{1}{|@{~}c@{~}|}{DANCE \cite{saito2020dance} } & 	\multicolumn{1}{@{~}c@{~}|}{3} & 	\multicolumn{1}{@{~}c@{~}|}{3} & 	\multicolumn{1}{@{~}c@{~}|}{fixed value depending on $|\sC_s|$ } 
\\
\multicolumn{1}{|@{~}c@{~}|}{PGL \cite{pgl-luo20b-icml20}} & 	\multicolumn{1}{@{~}c@{~}|}{3} & 	\multicolumn{1}{@{~}c@{~}|}{4} & 	\multicolumn{1}{@{~}c@{~}|}{reject a fixed portion of Target} \\
\multicolumn{1}{|@{~}c@{~}|}{MOSDANET \cite{rakshit2020multi}} & 	\multicolumn{1}{@{~}c@{~}|}{$4+|\sS|$} & 	\multicolumn{1}{@{~}c@{~}|}{2} & 	\multicolumn{1}{@{~}c@{~}|}{validated} \\
\hline
\multicolumn{1}{|@{~}c@{~}|}{\textbf{\OUR}} & 	\multicolumn{1}{@{~}c@{~}|}{1} & 	\multicolumn{1}{@{~}c@{~}|}{1} & 	\multicolumn{1}{@{~}c@{~}|}{self-paced, updates online while training} \\
\hline
\end{tabular}
}
\caption{Comparison with existing open-set and universal domain adaptation approaches. HPs indicate the hyperparameters, $|\sC_s|$ the number of source categories, $|\sS|$ is the number of source domains. Note that synthesizing new samples is a time-consuming operation and any validation procedure requires at least a dedicated per-dataset tuning. }\vspace{-5mm}
\label{tab:comparison}\vspace{-2mm}
\end{center}

\end{table}

\section{Related works}
\noindent\textbf{Domain Adaptation} 
A model trained and tested on data sharing the same label set but drawn from two different marginal distributions will inevitably show low performance.
\emph{Closed-Set domain adaptation} addresses this problem by increasing the invariance of the learned features over source and target domains. 
Several approaches focus on minimizing statistical metrics that reflect the distribution discrepancy \cite{Yan_2017_CVPR, Long_2015,Lee_2019_CVPR}. Others rely on adversarial learning \cite{ganin2016domain, NEURIPS2018_Long, tzeng2017adversarial}.
Recent strategies also exploit batch and feature normalization \cite{FMCarlucci_PAMI,batchnorm_ICLR,Xu_2019_ICCV} as well as self-supervision \cite{Carlucci_2019_CVPR,JiaolongXu2019} \cite{bucciTacklingPartialDomain2019,mitsuzumiGeneralizedDomainAdaptation2021} to learn robust cross-domain embeddings. 
A different stream of works investigates how to reduce the domain shift directly at pixel level via generative models which transfer the style of the source to the target and vice-versa \cite{russo17sbadagan,sankaranarayanan2017generate,Gong_2019_CVPR,NEURIPS2020_adain_forDA}.
When dealing with multiple sources, the extra challenge is in aligning all the domains among each other while producing a high discriminative feature space \cite{NEURIPS2020_3181d59d} . Source weighting techniques  
exploit 
 knowledge graphs and feature transferability measures evaluated once or multiple times over training \cite{peng2019moment, ECCV20_curriculumManager, ECCV20_learningToCombine}. 

Considering that the target is unlabeled, being sure that its semantic content perfectly matches that of the source is unrealistic. \emph{Open-Set domain adaptation} tackles target domains which include new unknown classes with respect to the source. After the definition of the problem in \cite{Busto_2017_ICCV}, a first group of works proposed various approaches to maximize the separation between known and unknown target samples while exploiting adversarial-based methods to align the known classes \cite{saito2018open,liu2019separate,feng2019attract}. Most recently, \cite{Pan_2020_CVPR} 
introduces 
a self-ensembling based method to minimize the model mismatch between the class assignment proposed by the source, and the inherent target cluster distribution. {ROS} \cite{bucci2020effectiveness} shows how to exploit the self-supervised rotation recognition task to deal with both these objectives. 
In \cite{kundu2020towards} a model is directly trained on the source with an extra set of negative samples produced via the suppression of class-specific feature maps activations.
PGL \cite{pgl-luo20b-icml20} exploits a graph neural network with episodic training to suppress the underlying conditional shift, while adversarial learning reduces the marginal shift between the source and target distributions.
The only published method dealing with multi-source Open-Set is {MOSDANET} \cite{rakshit2020multi} which adds a clustering objective over a standard supervised classification model to maximize the similarity among samples of the same class but different domains. Moreover, it exploits adversarial learning for domain adaptation: it has a tailored margin loss to penalize cases with a small difference in known and unknown prediction output, and finally it includes the potential target samples in the training procedure via pseudo-labeling.

The methods dealing with \emph{universal domain adaptation} cover a wide range of scenarios with private classes in source and/or target, including the Open-Set. In {DANCE} \cite{saito2020dance} a neighborhood clustering technique is integrated with the standard cross-entropy loss to learn the structure of the target, while an entropy-based score is used to align or reject the target samples. CMU \cite{fu2020learning} exploits a multi-classifier ensemble together with an unknown scoring function that combines entropy, confidence, and consistency measures.

\noindent\textbf{Contrastive Learning}
Lately, self-supervised learning methods have shown that, by relying only on unlabeled data, it is still possible to get classification performance similar to those of the supervised approaches
\cite{eccv2020_CMC,icml2020_CPC,simclr2020,MOCO_He_2020_CVPR}.
Contrastive learning builds over instance discrimination techniques \cite{Wu_2018_CVPR} (treating every instance as a class of its own), and aims at maximizing the agreement among multiple augmentations of the same sample, while pushing different instances far apart. 
Several methods have implemented this strategy by imposing the described constraints on the learned normalized embedding space. They differ in how positive and negative data pairs are sampled and stored:  among the most cited, SimCLR \cite{simclr2020} adopts a large batch size, while MoCo \cite{MOCO_He_2020_CVPR} maintains  a  momentum  encoder  and  a limited queue of previous samples. 
The effectiveness of the contrastive self-supervised learned embeddings is generally evaluated by using the pretext feature model as starting point for a downstream supervised task. However, more direct ways to incorporate supervision are currently attracting large attention \cite{NEURIPS2020_supclr,wei2020semantic} and show how view invariance and semantic knowledge can be combined to tackle novelty detection \cite{tack2020csi}, cross-domain generalization \cite{zhang2021unleashing} or few-shot classification \cite{majumder2021revisiting}.  Those approaches extend deep learning large margin models, demonstrating to be more robust across domains \cite{deepmetric_1,deepmetric_2,deepmetric_3}.
Current research is investigating ways to improve negative sampling in contrastive learning \cite{NEURIPS2020_debiased}, also proposing 
strategies to choose the best augmentation views  \cite{NEURIPS2020_goodviews,NEURIPS2020_demistifying}. 

\noindent\textbf{Learning on the Unit Hypersphere}
Fixed-norm representations have nice properties that support deep learning computational stability and their empirical success has been demonstrated over several tasks both within- and across-domains \cite{emnlp_XuD18, normface, Xu_2019_ICCV}. 
In particular, \cite{NEURIPS2019_hyperspherical_prototype} shows how setting class prototypes a priori on the unit hypersphere allows to free the output dimensionality from a constrained number of classes.
The uniform distribution of the data centers implies large margin separation among them and leaves space to include new categories while maintaining a highly discriminative embedding. 
A recent work has also highlighted how learning features uniformly distributed on the unit hypersphere with compact positive pairs is a crucial component of the success of contrastive learning \cite{icml_align_uniform}.

\section{Method}
\label{sec:method}
\label{sec:method}
\setlength{\belowdisplayskip}{5pt} \setlength{\belowdisplayshortskip}{5pt}
\setlength{\abovedisplayskip}{5pt} \setlength{\abovedisplayshortskip}{5pt}

To tackle multi-source Open Set domain adaptation we focus on building a robust, highly structured feature space with domain-aligned, compact, and well-separated class clusters, keeping \textit{unknown} target samples away from the centers. We obtain this effect by minimizing the supervised contrastive loss and paying attention to how data are fed to the model. In particular: (a) we design a domain and class balanced sampling strategy for mini-batch building, which allows to obtain a perfect class-wise alignment among the sources; (b) we add style transfer to the standard semantic-preserving transformations used to create sample pairs in contrastive learning, which provides a domain invariant feature embedding; (c) we refine source-target alignment by progressively including the target domain in the learning objective through self-training; (d) we tackle \textit{known-unknown} separation in the target domain by learning a self-paced threshold based on data distribution. We use this threshold both at inference time and when selecting \textit{known} target samples for self-training. In the following, after an introduction on the learning framework, we discuss all the listed points in detail. An overview of \OUR is illustrated in Figure \ref{fig:method} and summarized in Algorithm  \ref{alg:training} (see the supplementary material for the eval. procedure).

\begin{figure*}[ht!]
    \centering
\includegraphics[width=1\textwidth]{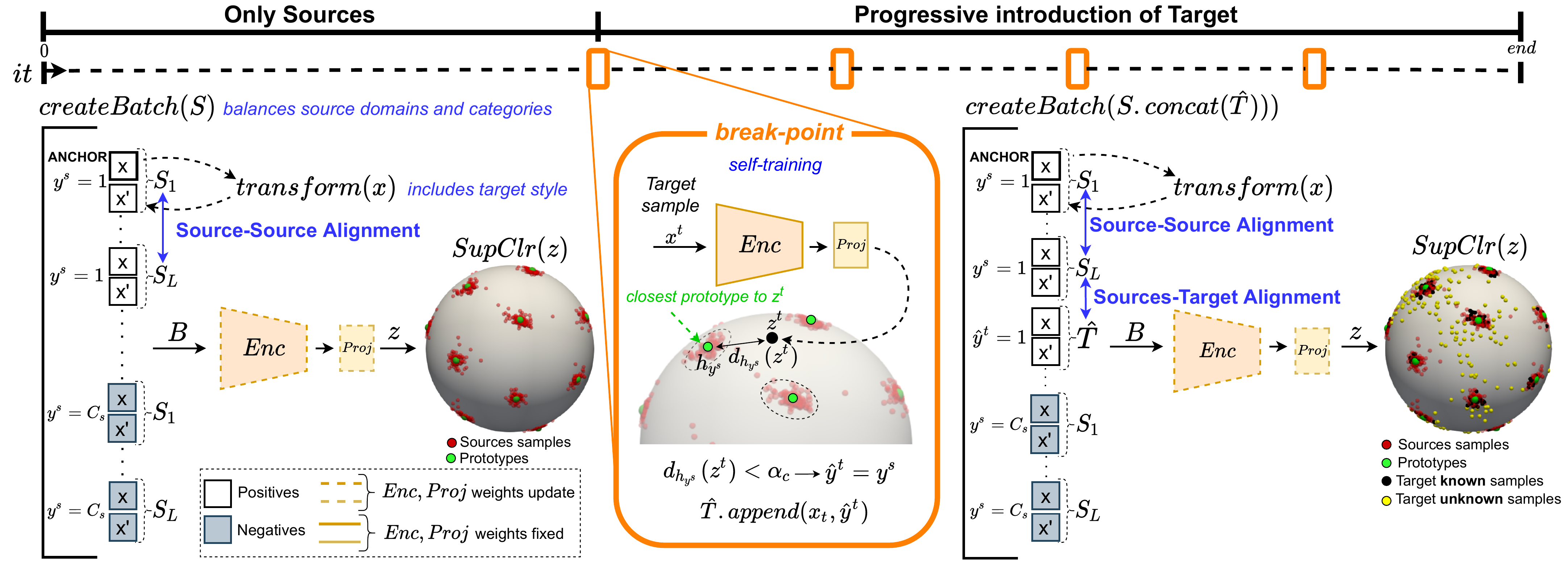}
    \caption{Schematic illustration of \OUR (best viewed in color). We use the same notation adopted in Algorithm \ref{alg:training}, please refer to it to follow the flow of the method.}
    \label{fig:method}\vspace{-3mm}
\end{figure*}
\setlength{\textfloatsep}{1pt}
\begin{algorithm}[tb]
\caption{\OUR training procedure}
\label{alg:training}
\begin{algorithmic}
\Require  $\{\bx^s,y^s\} \in \sS$, $\bx^t \in \sT$, $\alpha_m$, AdaIN model
\Ensure $Enc$ and $Proj$ 

\Procedure{transform}{$\bx$}
\State $styleAugment = random(True,False)$
\State $\bx' = randomCrop(\bx)$
\If {$styleAugment$}
\State \textbf{return} $styleTransf(\bx')$ {\color{blue} \Comment{target style}} 
\Else
\State \textbf{return} $grayScale(jitter((\bx')))$
\EndIf
\EndProcedure

\Procedure{createBatch}{$D$} {\color{blue} \Comment{$D$: set of domains} }
\State $batch = [\ ]$   {\color{blue} \Comment{balance domains and categories} }
\For{\textbf{each} $y^s$ \textbf{in} $\{1,...,|\sC_s|\}$}
\For{\textbf{each} $D_i$ \textbf{with} $i$ \textbf{in} $\{1,...,|D|\}$} 
\State $\bx'_{(y^s,D_i)} = transform(\bx_{(y^s,D_i)})$
\State $batch.append(\bx_{(y^s,D_i)},\bx'_{(y^s,D_i)})$
\EndFor
\EndFor
\State \textbf{return} $batch$ {\color{blue} \Comment{ $ len(batch) = |\sC_s| \times |D| \times 2 $}}
\EndProcedure

\Procedure{main()}{}
\State $ \hat{T} = [\ ] $
\For{$it$ \textbf{in} $range(0, end)$}
\If{$it$ \textbf{in} $break$-$points$ } 
\State $ \hat{T} = [\ ] $
\State $\alpha \leftarrow $ (Eq. \ref{eq:alpha}) ; $\alpha_c = \alpha_m \cdot \alpha $ 
\For{$x_t \textbf{ in } \sT$}
\State $\bz^t = Proj(Enc(\bx^t)$)

\State $ h_{y^s} \leftarrow \texttt{closest prototype to } \bz^t  $

\If{$d_{h_{y^s}}(\bz^t) < \alpha_c $ }  {\color{blue} \Comment{self-training} }
\State $ \hat{y}^t = y^{s}$ ; $ \hat{T}.append(\bx^t,\hat{y}^t) $

\EndIf
\EndFor
\EndIf

\State $B = createBatch(\sS.concat(\hat{T})) $
\State $ \bz = Proj(Enc(B)) $
\State $ loss = SupClr(\bz) $ (Eq. \ref{eq:supclr})
\State Update $Enc,Proj \leftarrow  \nabla loss$
\EndFor
\EndProcedure

\end{algorithmic}
\end{algorithm}

\noindent\textbf{Problem Framework}
In multi-source Open-Set domain adaptation we are given $L$ labeled source domains $\sS=\{\sS_1,\sS_2,\ldots,\sS_L\}$,  where ${\sS_i=\{\bx^{s_i}_j,y^{s_i}_j\}_{j=1}^{N^{s_i}}\sim p_i}$, and one unlabeled target domain ${\sT=\{\bx^t_j\}_{j=1}^{N^t}\sim q}$, all drawn from different data distributions $p_{i=1,\ldots,L}, q$. The sources share the same label set $y^s \in \sC_s $, and it holds $\sC_s\subset\sC_t$, thus the target \mbox{covers} $\sC_{t \setminus s}$ additional classes which are considered \textit{unknown}. Starting from this setup, the goal is to train a model on the source data, able to identify the label of each target sample, by either assigning it to one of the \textit{known} $\sC_s$ classes, or rejecting it as \textit{unknown}.
Given the different relatedness levels of the target with each of the available sources, reducing the domain shift while avoiding the risk of \emph{negative-transfer} may be difficult, especially when the \emph{openness} $\bbO=1-\frac{|\sC_s|}{|\sC_t|}$ increases.

\noindent\textbf{Contrastive Learning Formulation} 
In self-supervised contrastive learning \cite{simclr2020,MOCO_He_2020_CVPR}, two transformed views of every input image are propagated through a CNN network. The views are obtained via standard augmentation strategies as grayscale, random crop, and color jittering. 
For each sample $\{\bx^s_k,y^s_k\}$ in the double batch $B=\{k=1,\ldots,2K\}$
the features obtained via the encoder $Enc(\bx^s_k)$ enter the final contrastive head that further projects them to a normalized embedding, producing $\bz^s_k=Proj(Enc(\bx^s_k))$.
On the obtained hyperspherical space the samples are compared among each other:  the similarity between two augmented views of the same instance is maximized, while the similarity of two different instances is minimized.

When the image labels are available the sample comparison can be performed both instance-wise, as in the self-supervised case, and class-wise \cite{NEURIPS2020_supclr}: every samples of the same class $y^s_k$ are considered as positives, while the negative pairs are composed by samples of different classes. We indicate with $\nu(k)=B\setminus \{k\}$ the double batch without the \emph{anchor} sample of index $k$, and the positive pairs are $\pi(k) = \{k' \in \nu(k): y^s_{k'}=y^s_k\}$. Thus, the supervised contrastive loss is \cite{NEURIPS2020_supclr}:
\begin{equation}
    \sL_{SupClr} = \sum_{k=1}^{2K} \frac{-1}{|\pi(k)|} \sum_{k'\in \pi(k)} \log \frac{\exp(\sigma(\bz^s_k, \bz^s_{k'})/\tau)}{\sum\limits_{n\in \nu(k)} \exp(\sigma(\bz^s_k,\bz^s_{n})/\tau)}~,
    \label{eq:supclr}
\end{equation}
where $\tau\in \reals^+$ is the temperature, and $\sigma(\cdot, \cdot)$ is the cosine similarity.

\noindent\textbf{(a) \OUR Source-Source Class-Wise Domain Alignment}
The supervised contrastive loss aims at learning compact class clusters with large margins. This ability can be exploited to perform source-source class-wise domain alignment 
by composing each training mini-batch with samples coming from different visual domains. 
We evenly divide each batch to cover all the $|\sC_s|$ classes, and for each class, we select an equal number of samples from all the $L$ source domains. The loss in Eq. (\ref{eq:supclr}) does the rest, providing an embedding space where samples of the same class are concentrated in the same region regardless of the domain, while different classes are far apart from each other.

\noindent\textbf{(b) \OUR Source-Target Style Invariance}
The image transformations used in contrastive learning are meant to let the model focus on core semantic information while making it invariant to the irrelevant cues that they introduce. When dealing with data from different domains we desire a representation able to neglect major differences in visual appearance that go beyond mild grayscale or color jittering. This calls for dedicated semantic-preserving image transformations. 
We propose an augmentation based on style transfer as it is perfectly suitable for this goal: it does not affect the image content  while changing significantly the global image texture. 
In particular, we use the AdaIN \cite{huang2017adain} model trained jointly on source and target data  
in order to transfer the style from target images into source ones. As this augmentation is applied randomly, the loss function will explicitly compare original source images with target-like ones and learn to ignore the style difference. 

We highlight that our approach to obtain style invariance is safe from \textit{negative-transfer}.
This is one of the main issues in Open-Set domain adaptation due to the risk of aligning \textit{unknown} target categories to the \textit{known} ones of the source. All existing methods \cite{bucci2020effectiveness,fu2020learning,saito2020dance,you2019universal,liu2019separate} try to mitigate this problem by avoiding the inclusion of unknown samples or down-weighting them in the adaptation process. Thus, they are forced to identify the unknown samples before learning the domain invariant model by designing complex strategies. With style transfer, instead, we learn a domain agnostic representation since the beginning of the training process: it disregards the semantic content of the target so we can draw the style also from samples belonging to \textit{unknown} categories without incurring in \emph{negative-transfer}.

\noindent\textbf{(c) \OUR Adaptation Refinement via Self-Training}
In order to get a perfect source-target alignment, it would be enough to include the target data as an additional source domain while training the supervised contrastive model with the strategies described above.
Of course this is not possible as class labels are not available for target samples. 
However, once the model trained on source data and including target style invariance is robust enough, one could use it to produce pseudo-labels for target data by simply exploiting its predictions. 
We choose exactly this approach: 
after an initial source-only learning episode, we start progressively integrating the target samples in our learning objective, by passing through evaluation steps that we call \textit{self-training break-points}, which allow us to select confident \textit{known} target samples. Through this iterative technique, we propagate label knowledge from source to target data, improving the compactness of our class clusters while progressively leaving \textit{unknown} target data in low-density regions of the hyperspherical feature space.

\noindent\textbf{(d) \OUR Known-Unknown Separation and Classification on the Hypersphere}
The obtained embedding, with well clustered \emph{known} categories separated by large margins and \emph{unknown} samples in isolated areas, provides the ideal condition to perform distance-based classification. 
Differently from previous literature \cite{NEURIPS2020_supclr,tack2020csi} where the contrastive models were used only as pretext tasks and the projection head was later dropped in favor of a standard cross-entropy loss, we propose to stay on the hypersphere while delivering the final predictions.
We define the prototype of each source class $y^s$ by computing the corresponding feature average ${h_{y^s}=\frac{1}{N_{y^s}}\sum_{k\in y^s}\bz^s_{k}}\,,$ re-projected on the unit hypersphere. For any target sample $\bz^t$ we measure the cosine similarity to each source class prototype and we rescale it in $[0,1]$ to define the distance $d_{h_{y^s}}(\bz^t)=\{1-\sigma_{[0,1]}(\bz^t,h_{y^s})\}$ for $y^s\in\{1,\ldots,|\sC_s|\}\,,$ which is used as confidence measure for label assignment.

To decide whether a sample belongs to a \textit{known} category we need a threshold on the distance from the \textit{known} class prototypes. How to define such a threshold is a widely discussed problem in the Open Set literature, with many methods choosing values a priori and keeping them fixed while training \cite{fu2020learning,saito2020dance}. Instead, we propose to directly extract it from the observed data distribution, obtaining a value that changes online during the learning process. 
\begin{figure}[tb]
\centering
\includegraphics[width=7cm]{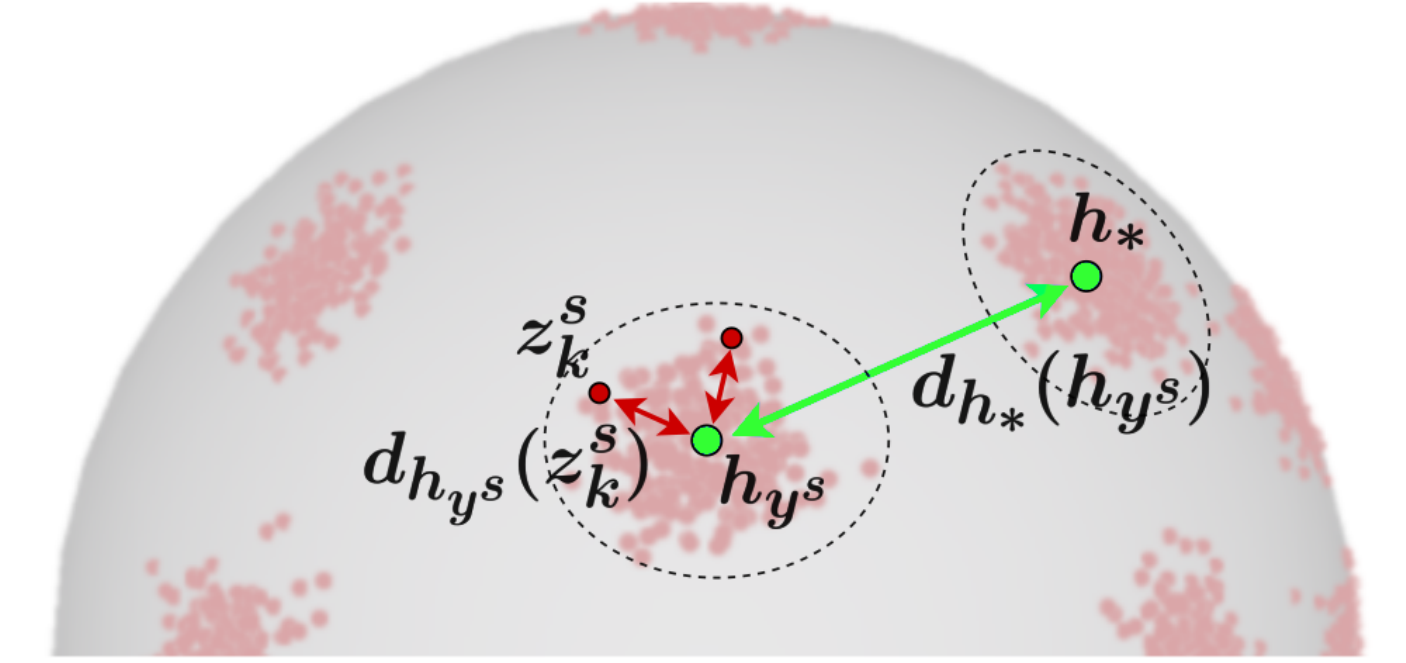}
    \caption{Illustration of the distances used to set the class prediction and the self-training procedure.}
    \label{fig:sphere}\vspace{3mm}
\end{figure}
\noindent Specifically we introduce two metrics to evaluate it: the \emph{class sparsity}:
\begin{equation}
    \theta = \frac{1}{|C_s|}\sum_{y^s\in C_s} d_{h_*}(h_{y^s})\,,
\end{equation}
where $h_*$ is the closest prototype to each $h_{y^s}$, and the \emph{class compactness}:
\begin{equation}
    \phi = \frac{1}{|C_s|}\sum_{y^s\in C_s}\left\{ \frac{1}{N_{y^s}} \sum_{k\in y^s} d_{h_{y^s}}(z_k^s)\right\}\,.
\end{equation}
In words, the former collects the prototype-to-prototype minimal distances and provides a measure of inter-class separation, while the latter evaluates whether the samples of each class are tight around the corresponding prototype (see Figure \ref{fig:sphere}). A dataset with a large number of categories, each with small intra-class variability, results in a feature scenario with high compactness but low sparsity, for which a low threshold is needed. On the other extreme, a dataset with a limited number of categories showing large intra-class variability corresponds to low compactness and high sparsity condition for which we can allow a higher threshold. We compute our threshold by:
\begin{equation}
    \alpha= \phi \cdot \left[ log \left( \cfrac{\theta}{2\phi}\right) +1 \right]\,,
    \label{eq:alpha}
\end{equation}
where ${\theta}/{2\phi}$ estimates the average ratio between the distance of two adjacent prototypes and the radii of the respective clusters.
The use of the threshold at inference time is straightforward:
\begin{equation}
    \hat{y}^t = 
    \begin{cases}
\argmin_{y^s}(d_{h_{y^s}}(\bz^t)) & \mbox{if } \min{_{y^s}(d_{h_{y^s}}(\bz^t))}< \alpha\\
\texttt{unknown} & \mbox{if } \min{_{y^s}(d_{h_{y^s}}(\bz^t))}\geq \alpha\,.\\
    \end{cases}
\end{equation}
We exploit this threshold also for the {self-training {break-points}} described before. Only in this phase we are particularly cautious
, thus we include a multiplier $\alpha_m$  
that allows us to keep a more conservative threshold: $\alpha_c = \alpha_m \cdot \alpha$. This multiplier can be kept fixed to $0.5$ and it is the only hyperparameter of \OUR.

\section{Experiments}
We  implemented  \OUR with a  ResNet-50 \cite{resnet}  backbone  and  two  fully  connected  layers  which  define the  contrastive  head. All the implementation details as well as the Pytorch code are provided in the supplementary material.

\noindent\textbf{Datasets} 
We evaluate our approach on three image classification benchmarks, following the same setting used in \cite{rakshit2020multi}, with one domain considered in turn as target. 
{\textbf{Office31}} \cite{saenko2010adapting} comprises three domains: Webcam ({W}), Dslr ({D}) and Amazon ({A}) each containing 31 object categories. We set as known the first 20 classes in alphabetic order, while the remaining 11 are unknown. 
{\textbf{Office-Home}} \cite{venkateswara2017deep} is made by four domains: Art ({Ar}), Clipart ({Cl}), Product ({Pr}), RealWorld ({Rw}) with 65 classes. The first 45 categories in alphabetic order are known, and the remaining 20 are unknown. 
{\textbf{DomainNet}} \cite{peng2019moment} is a more challenging testbed than the previous ones.  It contains six domains and 345 classes. As in \cite{rakshit2020multi}, we consider Infograph ({I}), Painting ({P}), Sketch ({S}) and Clipart ({C}), selecting randomly 50 samples per class or using all the images in case of lower cardinality.  The first 100 classes in alphabetic order are known, while the remaining 245 are unknown.

\noindent\textbf{Results}
We compare \OUR with several state-of-the-art baselines proposed for single-source Open-Set  (Inheritable \cite{kundu2020towards}, ROS \cite{bucci2020effectiveness}, PGL \cite{pgl-luo20b-icml20}), multi-source Open-Set (MOSDANET \cite{rakshit2020multi}) and universal domain adaptation (CMU \cite{fu2020learning}, DANCE \cite{saito2020dance}). 
We use the code provided by the authors\footnote{For all the baseline methods the implementations are publicly available, with the only exception of MOSDANET \cite{rakshit2020multi} for which we obtained the code via private communications with the authors.}, and for all the methods that do not specify how to manage multiple sources, we apply the  \emph{Source Combine} strategy  \cite{peng2019moment} that considers the union of all the source data in a single domain.  
We adopt the $HOS$ metric, defined in \cite{bucci2020effectiveness,fu2020learning} for a fair evaluation of Open-Set methods: it is the harmonic mean between the average class accuracy over the known classes \emph{OS$^*$} and the accuracy over the unknown class \emph{UNK}: $HOS= 2 \frac{\text{{OS}$^*$} \times \text{{UNK}}}{\text{{OS$^*$}} + \text{{UNK}}}$.

Table \ref{tab:sota} collects the obtained results showing how \OUR outperforms all the baselines. The gain of \OUR with respect to the best competitor ROS, varies from 1.9\% points on OfficeHome, up to 10.8\% on DomainNet. Besides being simpler than the reference approaches, \OUR shows to be robust to the significantly different scenarios covered by the three datasets in terms of number of shared and private classes, as well as nature and extent of the domain gaps. These peculiarities make \OUR the most suitable approach in a variety of real-world applications. 

We also benchmark \OUR with the best competitor ROS in terms of the $AUROC$ (Area under the Receiver Operating Characteristic curve) metric, which has the advantage of being threshold-independent. In our case, the \emph{normality score} used to evaluate whether a sample is \emph{known} or \emph{unknown} is its distance from the nearest source class prototype, while ROS exploits a combination of entropy and probability output of an auxiliary rotation recognition classifier. Even in this case, \OUR outperforms ROS, reflecting what is already observed in terms of HOS. This also confirms how well \emph{known} and \emph{unknown} samples are separated in the learned hyperspherical embedding space.

\begin{table*}[t]
\centering
\resizebox{\textwidth}{!}{
\begin{tabular}{c|ccccc|}
\cline{2-6}
 & \multicolumn{2}{|c}{\multirow{2}{*}{Method}}  & \multicolumn{3}{|c|}{\textbf{DomainNet}} \\
\cline{4-6}
& & & \multicolumn{1}{|c|}{ $\rightarrow$ S } & \multicolumn{1}{c|}{ $\rightarrow$ C } & \multicolumn{1}{c|}{\textbf{Avg.}}\\
\hline
\multicolumn{1}{|c}{\multirow{7}{*}{\rotatebox{90}{HOS}}} & \multicolumn{1}{|c}{\multirow{5}{*}{Source Combine }} 
 & \multicolumn{1}{c|}{Inheritable  \cite{kundu2020towards}} & 	\multicolumn{1}{c|}{34.8}  & \multicolumn{1}{c|}{44.0} & 39.4 \\

\multicolumn{1}{|c|}{} & &  \multicolumn{1}{c|}{ROS \cite{bucci2020effectiveness}}  &	\multicolumn{1}{c|}{44.5} &	\multicolumn{1}{c|}{52.4}  & 48.5 \\
\multicolumn{1}{|c|}{} & &  \multicolumn{1}{c|}{CMU \cite{fu2020learning}} & \multicolumn{1}{c|}{38.1}  & \multicolumn{1}{c|}{35.5} & 36.8\\  
\multicolumn{1}{|c|}{} & &  \multicolumn{1}{c|}{DANCE \cite{saito2020dance}}  & \multicolumn{1}{c|}{30.0}  & \multicolumn{1}{c|}{37.6}  & 33.8 \\ 
\multicolumn{1}{|c|}{} & &  \multicolumn{1}{c|}{PGL \cite{pgl-luo20b-icml20}}  & \multicolumn{1}{c|}{18.5}  & \multicolumn{1}{c|}{19.4}  & 19.0 \\ 
\cline{2-6}
\multicolumn{1}{|c|}{} & \multicolumn{1}{|c}{\multirow{2}{*}{Multi-Source }} & \multicolumn{1}{c|}{MOSDANET \cite{rakshit2020multi}}  & 	\multicolumn{1}{c|}{40.0}  & \multicolumn{1}{c|}{39.3}  & 39.6\\
\multicolumn{1}{|c|}{} & & \multicolumn{1}{c|}{\textbf{\OUR}} & \multicolumn{1}{c|}{\textbf{57.5}}  &	\multicolumn{1}{c|}{\textbf{61.0}}  &	\textbf{59.3}\\
\hline\hline
\multicolumn{1}{|c}{\multirow{2}{*}{\rotatebox{90}{\scriptsize{AUROC}}}} & \multicolumn{1}{|c}{Source Combine} 
 & \multicolumn{1}{c|}{ROS  \cite{bucci2020effectiveness}} & 	\multicolumn{1}{c|}{63.9}  & \multicolumn{1}{c|}{68.0} & 66.0 \\
 \multicolumn{1}{|c}{\multirow{2}{*}{}} & \multicolumn{1}{|c}{Multi-Source}  & \multicolumn{1}{c|}{\textbf{\OUR}} & 	\multicolumn{1}{c|}{\textbf{71.9}}  & \multicolumn{1}{c|}{\textbf{75.8}} & \textbf{73.9} \\
 \hline
\end{tabular}
\begin{tabular}{|cccc|}
\hline
\multicolumn{4}{|c|}{\textbf{Office31}}        \\
\hline
\multicolumn{1}{|c|}{$\rightarrow$ W } & \multicolumn{1}{c|}{ $\rightarrow$ D } & \multicolumn{1}{c|}{$\rightarrow$ A } &
\multicolumn{1}{c|}{\textbf{Avg.}}\\
\hline
\multicolumn{1}{|c|}{76.6}  & \multicolumn{1}{c|}{79.5}  & \multicolumn{1}{c|}{70.0}  & 75.4 \\
\multicolumn{1}{|c|}{81.8} & \multicolumn{1}{c|}{80.1} &	\multicolumn{1}{c|}{64.7} &	75.5 \\
\multicolumn{1}{|c|}{61.4}  &	\multicolumn{1}{c|}{64.0} &	\multicolumn{1}{c|}{56.4}  & 60.6 \\
\multicolumn{1}{|c|}{38.5}  &	\multicolumn{1}{c|}{59.7}   &	\multicolumn{1}{c|}{58.0}  &	52.0  \\ 
\multicolumn{1}{|c|}{43.3}  &	\multicolumn{1}{c|}{37.7}   &	\multicolumn{1}{c|}{35.6}  &	38.9  \\ 
\hline
\multicolumn{1}{|c|}{60.5}  &	\multicolumn{1}{c|}{71.5}  &	\multicolumn{1}{c|}{\textbf{73.9}}  & 68.6\\
\multicolumn{1}{|c|}{\textbf{90.2}}  &	\multicolumn{1}{c|}{\textbf{89.9}}  &	\multicolumn{1}{c|}{60.8}  &		\textbf{80.3}\\
\hline \hline
\multicolumn{1}{|c|}{{93.9}}  &	\multicolumn{1}{c|}{{95.2}}  &	\multicolumn{1}{c|}{\textbf{73.5}}  &		{87.5}\\
\multicolumn{1}{|c|}{\textbf{96.9}}  &	\multicolumn{1}{c|}{\textbf{96.1}}  &	\multicolumn{1}{c|}{71.0}  &		\textbf{88.0}\\
\hline
\end{tabular}
\begin{tabular}{|ccccc|}
\hline
\multicolumn{5}{|c|}{\textbf{Office-Home}}        \\
\hline
\multicolumn{1}{|c|}{ $\rightarrow$ Rw } & \multicolumn{1}{c|}{ $\rightarrow$ Cl } & \multicolumn{1}{c|}{ $\rightarrow$ Ar } & \multicolumn{1}{c|}{ $\rightarrow$ Pr } & 
\multicolumn{1}{c|}{\textbf{Avg.}}\\
\hline
\multicolumn{1}{|c|}{63.2}  &	\multicolumn{1}{c|}{52.6}  &	\multicolumn{1}{c|}{48.7}  &	\multicolumn{1}{c|}{60.7}  &	56.3\\
\multicolumn{1}{|c|}{\textbf{73.0}}   &	\multicolumn{1}{c|}{57.3}   &	\multicolumn{1}{c|}{61.6}   &	\multicolumn{1}{c|}{69.1} &	65.3\\
\multicolumn{1}{|c|}{70.8} &			\multicolumn{1}{c|}{50.0}  &		\multicolumn{1}{c|}{58.1}  &		\multicolumn{1}{c|}{69.3}  &	62.1 \\ 
\multicolumn{1}{|c|}{12.4} &	  	\multicolumn{1}{c|}{16.1}  &	\multicolumn{1}{c|}{18.6}  &	\multicolumn{1}{c|}{22.9}  & 17.5 \\
\multicolumn{1}{|c|}{40.0} &	  	\multicolumn{1}{c|}{31.5}  &	\multicolumn{1}{c|}{31.8}  &	\multicolumn{1}{c|}{42.2}  & 36.4 \\
 \hline
\multicolumn{1}{|c|}{65.0} &  \multicolumn{1}{c|}{51.1}  &	\multicolumn{1}{c|}{54.3}  &	\multicolumn{1}{c|}{65.9}  &	59.1\\
\multicolumn{1}{|c|}{71.0}  &		\multicolumn{1}{c|}{\textbf{64.6}}  &		\multicolumn{1}{c|}{\textbf{62.2}}  &	\multicolumn{1}{c|}{\textbf{71.1}} &	\textbf{67.2} \\
\hline\hline
\multicolumn{1}{|c|}{80.8} &  \multicolumn{1}{c|}{69.6}  &	\multicolumn{1}{c|}{73.7}  &	\multicolumn{1}{c|}{79.4}  &	75.9\\
\multicolumn{1}{|c|}{\textbf{81.1}}  &		\multicolumn{1}{c|}{\textbf{76.4}}  &		\multicolumn{1}{c|}{\textbf{75.3}}  &	\multicolumn{1}{c|}{\textbf{79.6}} &	\textbf{78.1} \\
\hline
\end{tabular}
}
\caption{Results averaged over three runs for each method on the  DomainNet, Office31, and Office-Home datasets. }
\label{tab:sota} \vspace{-2mm}
\end{table*}

\begin{figure}[t!]
\hspace{-4mm}
\resizebox{0.52\textwidth}{!}{
\begin{tabular}{@{}c@{}c}
\includegraphics[width=0.26\textheight]{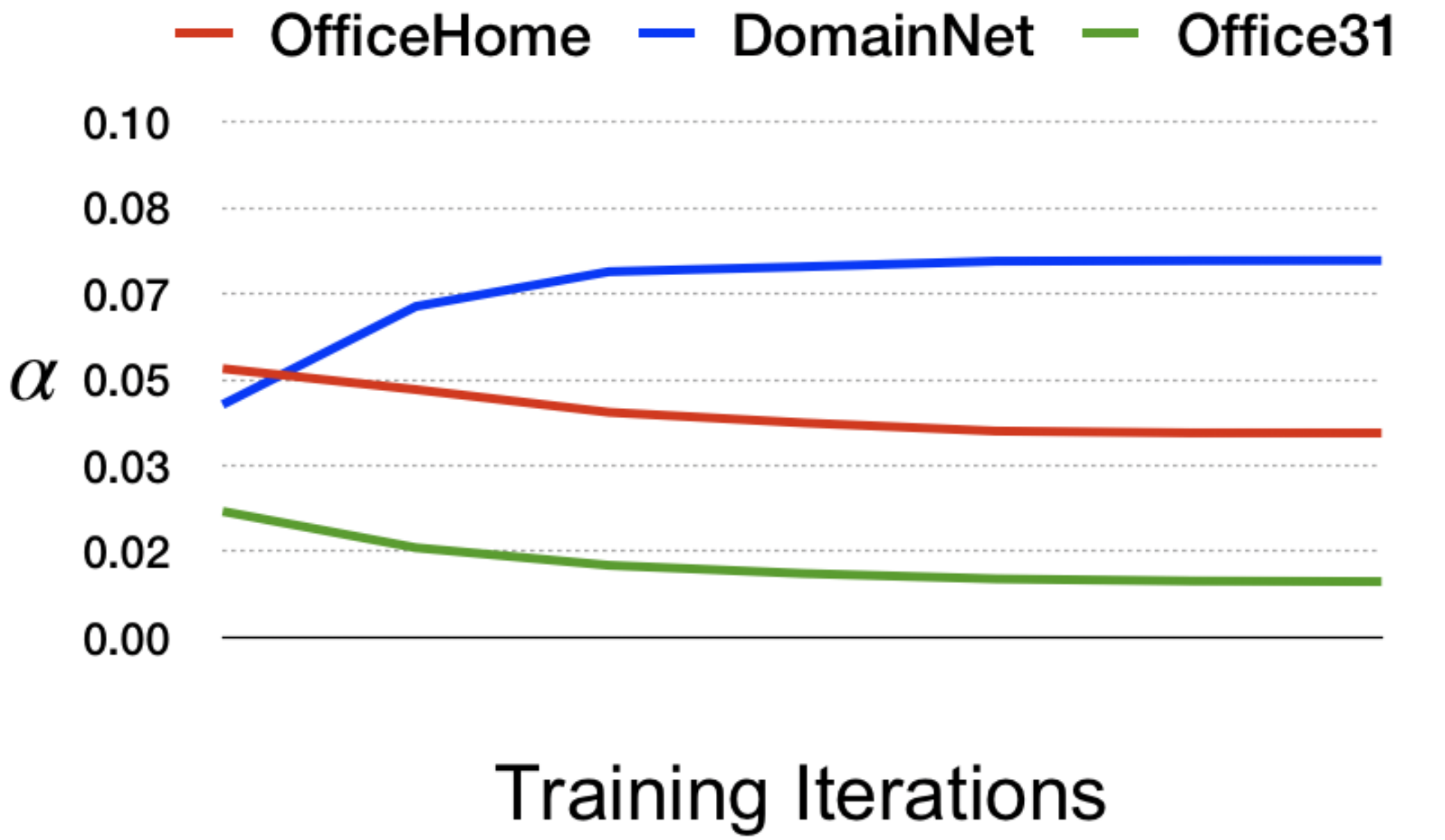} &
\includegraphics[width=0.22\textheight]{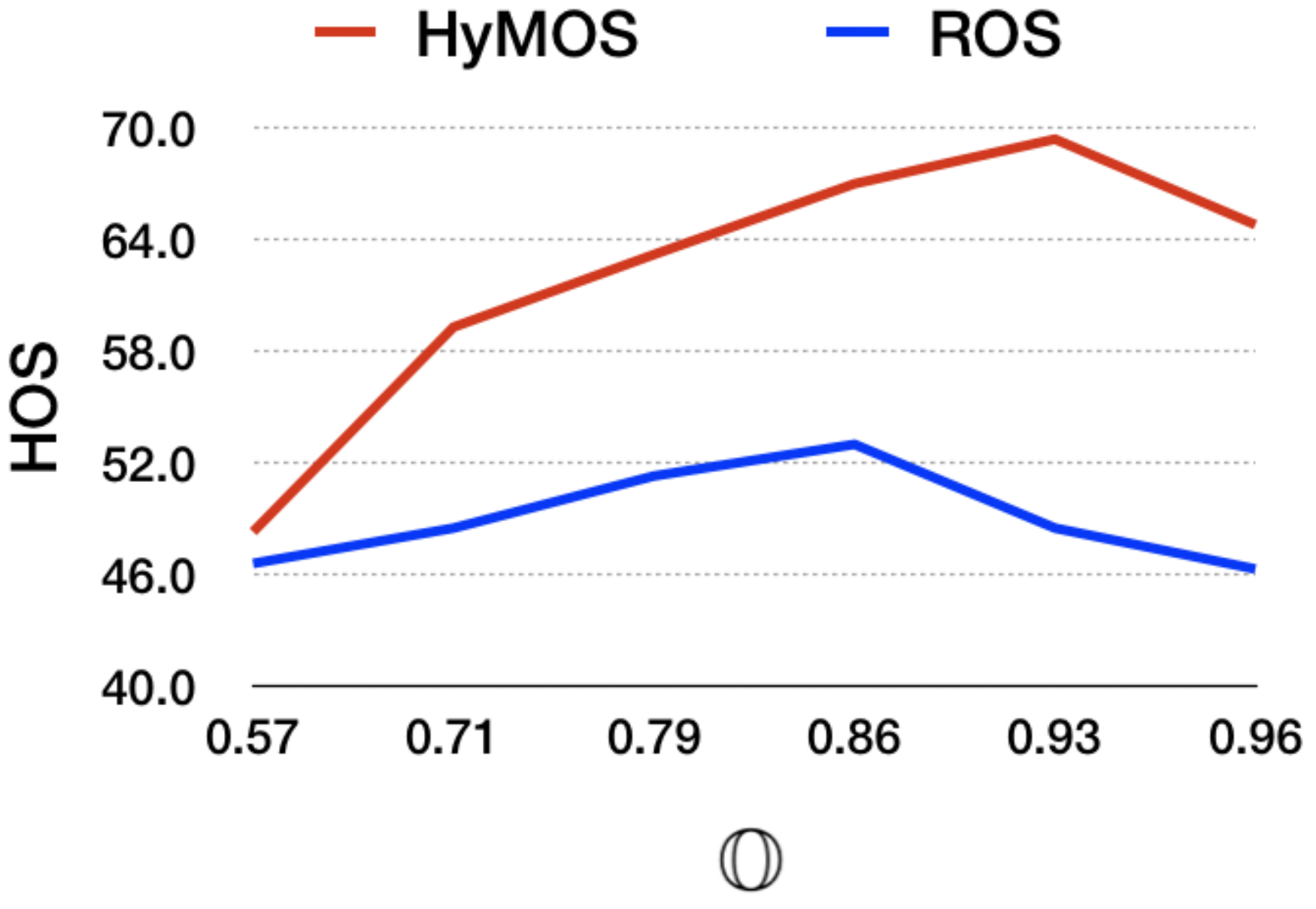} 
\end{tabular}
}\vspace{-2mm}
\caption{Left: analysis on the dynamic threshold $\alpha$ at different training iterations. Right: performance of \OUR and ROS \cite{bucci2020effectiveness} at different openness ($\mathbb{O}$) levels.}
\label{fig:iterations}\vspace{-2mm}
\end{figure}

\noindent\textbf{Analysis on the Threshold}
For \OUR we designed a self-paced procedure that learns the dynamic threshold $\alpha$ from the data distribution. 
Figure \ref{fig:iterations} (left) provides an overview of $\alpha$ at different training iterations
: for Office31 and Office-Home the threshold decreases over time while for DomainNet it increases. These variations evidence how the data clusters move: as the training proceeds they become more compact and the reciprocal distance increases towards a more uniform class distribution on the hypersphere. For DomainNet the second event occurs faster than the first: this trend is correlated with the class cardinality  which is higher with respect to that of  other datasets. In all the cases, the threshold converges to a stable value.  

The $\alpha_m$ multiplier used at training time to compute a conservative threshold is the only hyperparameter of \OUR: by modifying it one could decide to favor recognition of \emph{known} classes at the expense of a lower \emph{unknown} recognition during training. The results in Table \ref{tab:alphamultiplier} show that $\alpha_m$=0.5 is a safe choice regardless of the dataset. Moreover, by tuning this multiplier, the HOS performance of \OUR remains always competitive with  ROS, and can even increase as in the case of DomainNet for $\alpha_m$=1.

\begin{table}[tb]
\begin{center}
\begin{tabular}{|@{~}c@{~}c@{~}c@{~~~}c@{~~~}c@{~~~}}
\hline
\multicolumn{2}{|@{~}c@{~}|}{Method} &  \multicolumn{1}{|@{~}c@{~}|}{\textbf{DomainNet}} &  \multicolumn{1}{|@{~}c@{~}}{\textbf{Office31}} &  \multicolumn{1}{|@{~}c@{~}|}{\textbf{Office-Home}} \\
\hline
{\multirow{4}{*}{\OUR}} &
\multicolumn{1}{@{~}c@{~}|}{$\alpha_{m}$ = 0.3} & \multicolumn{1}{@{~}c@{~}|}{55.1} &	\multicolumn{1}{@{~}c@{~}|}{79.2}&	\multicolumn{1}{@{~}c@{~}|}{65.8} \\
&\multicolumn{1}{@{~}c@{~}|}{{\underline{$\alpha_{m}$ = 0.5}}} & 	\multicolumn{1}{@{~}c@{~}|}{59.3} & 	\multicolumn{1}{@{~}c@{~}|}{\textbf{80.3}} & 	\multicolumn{1}{@{~}c@{~}|}{\textbf{67.2}} \\
&\multicolumn{1}{@{~}c@{~}|}{$\alpha_{m}$ = 0.7} & 	\multicolumn{1}{@{~}c@{~}|}{60.8} & 	\multicolumn{1}{@{~}c@{~}|}{78.2} & 	\multicolumn{1}{@{~}c@{~}|}{66.8} \\
&\multicolumn{1}{@{~}c@{~}|}{$\alpha_{m}$ = 1.0} & 	\multicolumn{1}{@{~}c@{~}|}{\textbf{61.4}}  &  \multicolumn{1}{@{~}c@{~}|}{74.1} & 	\multicolumn{1}{@{~}c@{~}|}{65.8} \\
\hline
\multicolumn{2}{|@{~}c@{~}|}{ROS \cite{bucci2020effectiveness}} & 	\multicolumn{1}{@{~}c@{~}|}{48.5}  & 
\multicolumn{1}{@{~}c@{~}|}{75.7}  & \multicolumn{1}{@{~}c@{~}|}{65.3} \\
\hline
\end{tabular}
\caption{Average performance (HOS) when changing the train-time multiplier $\alpha_m$ to the self-paced threshold $\alpha$.}
\label{tab:alphamultiplier}
\end{center}
\end{table}


\noindent\textbf{Increasing the Openness Level}
In real-world conditions, it is difficult to have direct control over the number of \emph{unknown} classes in the unlabeled target, and it is natural to expect more \emph{unknown} categories than \emph{known} ones. To study how \OUR reacts at different openness levels, we consider the DomainNet dataset and exploit its large class cardinality.
The plot in Figure \ref{fig:iterations} (right) shows the HOS accuracy of \OUR and how it outperforms its best competitor ROS at different openness values $\mathbb{O}\in\{0.5,1\}$.


\section{Ablation Analysis}

\begin{table}
\resizebox{1.01\linewidth}{!}{
\begin{tabular}{c c c c c c|}
\hline
\multicolumn{1}{|c|}{\multirow{2}{*}{Method}} &\multicolumn{5}{c|}{\textbf{Office-Home}}        \\
\cline{2-6}
\multicolumn{1}{|c|}{}&  \multicolumn{1}{c|}{$\rightarrow$ Rw } & \multicolumn{1}{c|}{ $\rightarrow$ Cl } & \multicolumn{1}{c|}{ $\rightarrow$ Ar } & \multicolumn{1}{c|}{ $\rightarrow$ Pr } & 
\multicolumn{1}{c|}{\textbf{Avg.}}\\
 \hline
    \multicolumn{1}{|c|}{\textbf{\OUR}}  &		\multicolumn{1}{c|}{71.0}  &	\multicolumn{1}{c|}{64.6}  &	\multicolumn{1}{c|}{62.2} &		\multicolumn{1}{c|}{71.1}  &	\textbf{67.2} \\
       \multicolumn{1}{|c|}{w/o Source Balance}  &	\multicolumn{1}{c|}{69.2}  &	\multicolumn{1}{c|}{58.4}  &	\multicolumn{1}{c|}{60.6} &	\multicolumn{1}{c|}{70.2}  & 64.6\\

    \multicolumn{1}{|c|}{Style Tr. Target Known (Oracle)}  &	\multicolumn{1}{c|}{70.7}  &	\multicolumn{1}{c|}{63.7}  &	\multicolumn{1}{c|}{62.5}  &	\multicolumn{1}{c|}{71.2} &	67.0 \\
   \multicolumn{1}{|c|}{w/o Style Transfer}  &	\multicolumn{1}{c|}{69.5}  &	\multicolumn{1}{c|}{56.4}  &	\multicolumn{1}{c|}{60.0}  &	\multicolumn{1}{c|}{68.3}  &	63.6 \\
   \multicolumn{1}{|c|}{w/o Self-Training} 	&	\multicolumn{1}{c|}{72.2}  &	\multicolumn{1}{c|}{55.0}  &	\multicolumn{1}{c|}{58.6}  &	\multicolumn{1}{c|}{71.5}  & 	64.3\\
\hline\hline
\multicolumn{1}{|c|}{Improved Cross-Entropy}  &	\multicolumn{1}{c|}{61.5}  &	\multicolumn{1}{c|}{61.2}  &	\multicolumn{1}{c|}{58.1}  &	\multicolumn{1}{c|}{57.1}  &	59.5 \\
 \hline\hline
   \multicolumn{1}{|c|}{ROS \cite{bucci2020effectiveness}} &	\multicolumn{1}{c|}{73.0}  &	\multicolumn{1}{c|}{57.3}  &	\multicolumn{1}{c|}{61.6}  &	\multicolumn{1}{c|}{69.1} &	65.3\\
\multicolumn{1}{|c|}{+ Source Balance} &		\multicolumn{1}{c|}{75.2} &	\multicolumn{1}{c|}{55.5} &	\multicolumn{1}{c|}{62.6} &		\multicolumn{1}{c|}{66.9} &	65.0\\
\multicolumn{1}{|c|}{+ Style Transfer }  &	\multicolumn{1}{c|}{62.6} &	\multicolumn{1}{c|}{46.3}  &	\multicolumn{1}{c|}{52.0}  &	\multicolumn{1}{c|}{60.1}  &	55.2\\
\multicolumn{1}{|c|}{+ Self-Training} &	\multicolumn{1}{c|}{69.6} &	\multicolumn{1}{c|}{59.1} &		\multicolumn{1}{c|}{61.5} &	\multicolumn{1}{c|}{60.5} &	62.7\\
\multicolumn{1}{|c|}{+ S. Balance, Style Tr., Self-Train.} &	\multicolumn{1}{c|}{62.0} &	\multicolumn{1}{c|}{40.4} &	\multicolumn{1}{c|}{52.2} &	\multicolumn{1}{c|}{62.4} &	54.3 \\
 \hline
\end{tabular}
}
\caption{Ablation Study, HOS results.}\vspace{3mm}
\label{tab:ablation}
\end{table}

\begin{table*}[h]
\resizebox{\textwidth}{!}{
\begin{tabular}[t]{|c c c c c c c c|}
\hline
\multicolumn{8}{|c|}{\textbf{Multi-Source Closed-Set}}        \\
\hline
Method & \multicolumn{1}{|c|}{$\rightarrow$ clp } 
& \multicolumn{1}{|c|}{$\rightarrow$ inf } 
& \multicolumn{1}{|c|}{$\rightarrow$ pnt  } 
& \multicolumn{1}{|c|}{$\rightarrow$ qdr } 
& \multicolumn{1}{|c|}{$\rightarrow$ rel  } 
& \multicolumn{1}{|c|}{$\rightarrow$ skt }  &
\multicolumn{1}{c|}{\textbf{Avg.}} \\
 \hline
 \multicolumn{1}{|c|}{Source Only \cite{li2021dynamic}}  &		\multicolumn{1}{c|}{52.1} &	\multicolumn{1}{c|}{23.4}  &	\multicolumn{1}{c|}{47.7}  &	\multicolumn{1}{c|}{13.0}  &	\multicolumn{1}{c|}{60.7}  &	\multicolumn{1}{c|}{46.5} &	40.6   \\
\multicolumn{1}{|c|}{LtC-MSDA \cite{wang2020learning}}  &		\multicolumn{1}{c|}{63.1} &	\multicolumn{1}{c|}{28.7}  &	\multicolumn{1}{c|}{56.1}  &	\multicolumn{1}{c|}{16.3}  &	\multicolumn{1}{c|}{66.1}  &	\multicolumn{1}{c|}{53.8} &	47.4  \\
\multicolumn{1}{|c|}{DRT \cite{li2021dynamic}}  &		\multicolumn{1}{c|}{71.0} &	\multicolumn{1}{c|}{31.6}  &	\multicolumn{1}{c|}{\textbf{61.0}}  &	\multicolumn{1}{c|}{12.3}  &	\multicolumn{1}{c|}{71.4}  &	\multicolumn{1}{c|}{60.7} &	51.3 \\
\multicolumn{1}{|c|}{\textbf{\OUR}}  &		\multicolumn{1}{c|}{\textbf{71.5}} &	\multicolumn{1}{c|}{\textbf{41.8}}  &	\multicolumn{1}{c|}{60.8}  &	\multicolumn{1}{c|}{\textbf{34.5}}  &	\multicolumn{1}{c|}{\textbf{74.2}}  &	\multicolumn{1}{c|}{\textbf{66.6}} &	\textbf{58.2}   \\
\hline
\end{tabular}
\begin{tabular}[t]{|c c c c|}
\hline
\multicolumn{4}{|c|}{\textbf{Multi-Source Universal}}        \\
\hline
Method &  \multicolumn{1}{|c|}{$\rightarrow$ S } & \multicolumn{1}{c|}{$\rightarrow$ C } & \multicolumn{1}{c|}{\textbf{Avg.}}\\
 \hline
 \multicolumn{1}{|c|}{CMU \cite{fu2020learning}}  &		\multicolumn{1}{c|}{38.9}  & \multicolumn{1}{c|}{31.2}  &  35.1 \\
 \multicolumn{1}{|c|}{DANCE \cite{saito2020dance}}  & 		\multicolumn{1}{c|}{44.5}  & \multicolumn{1}{c|}{49.9}  &  47.2 \\
 \multicolumn{1}{|c|}{ROS \cite{bucci2020effectiveness}}  & 		\multicolumn{1}{c|}{39.7}  & \multicolumn{1}{c|}{46.0}  & 42.9  \\
 \multicolumn{1}{|c|}{\textbf{\OUR}}  & 
    \multicolumn{1}{c|}{\textbf{54.6}}  & 
    \multicolumn{1}{c|}{\textbf{57.1}}  & 
    \textbf{55.9}\\
 \hline
\end{tabular}
}
\vspace{-2mm}\caption{Multi-Source Closed-Set (Accuracy) and Universal Domain Adaptation (HOS) performance on DomainNet.}
\label{tab:Closed-Set}\vspace{-3mm}
\end{table*}

We designed \OUR to be straightforward while keeping in mind all the challenges of multi-source Open-Set domain adaptation. In the following we focus on each of them, providing a detailed ablation that sheds light on the inner functioning of our method. The results are in Table \ref{tab:ablation}.

\noindent\textbf{Source-Source Alignment}
Reducing the domain shift among the available sources improves model generalization. 
This aspect is largely discussed in multi-source Closed-Set domain adaptation literature  \cite{ECCV20_curriculumManager,ECCV20_learningToCombine}. A dedicated source alignment component is also included in the only existing multi-source Open-Set method MOSDANET.

\OUR obtains cross-source adaptation by combining the supervised contrastive learning loss with an accurately designed batch sampling strategy: each training mini-batch contains one sample for each class and for each domain. 
The supervised contrastive loss provides a strong class-wise alignment by pulling together samples of the same class and pushing away samples of different classes regardless of the domain. 
\OUR shows a gain in performance of 2.6\% over its version without this balancing (row \emph{w/o Source Balance}). 

\noindent\textbf{Source-Target Adaptation}
In \OUR, both the style transfer augmentation and the auto-regulated self-training procedure contribute to aligning source and target without incurring the risk of \emph{negative-transfer}. 
By adding target style transfer as one of the source augmentations we push the model to focus on domain agnostic visual characteristics without involving semantic content from the target. To evaluate the effect of this addition we present two ablation cases. 
We compare our method with an Oracle version where the target style is extracted only from \emph{known} categories (line \emph{Style Tr. Target Known (Oracle)}), and we conclude that \OUR is not harmed when using the whole target for this adaptation step.
Moreover, we deactivate style transfer (row \emph{w/o Style Transfer}) causing a performance drop of 3.6\%, which shows its important role in \OUR.

Finally, a strong feature-level class-wise source-target alignment is obtained thanks to the self-training procedure, which selects confident target known samples (closest to the source class prototypes) and includes them in the learning objective. The gain of \OUR with respect to its version without this strategy is 2.9\% (row \emph{w/o Self-Training}).

\noindent\textbf{Comparison with an Improved Cross-Entropy Baseline}
Source balance, style transfer, and self-training appear as simple strategies that can be combined with any supervised learning model to improve its effectiveness in the multi-source Open-Set scenario. Still, we state that leveraging on supervised contrastive learning and its related hyperspherical embedding is crucial for the task at hand. 
To support our claim we substitute the contrastive loss of \OUR with the standard cross-entropy loss.
The row \emph{Improved cross-entropy} reports the obtained results, showing that this baseline approach is significantly worse than \OUR.

\noindent\textbf{Comparison with an improved version of ROS~\cite{bucci2020effectiveness}}
We also enriched our best competitor ROS with source balancing, style transfer, and self-training. 

In the bottom part of Table \ref{tab:ablation}, the \emph{+ Source Balance} row indicates that organizing the training data batches so that they contain a balanced set of categories and source domains does not provide an improvement with respect to the standard version of ROS. The  source-to-source alignment visible for \OUR does not appear here: indeed
the cross-entropy loss does not induce the same inherent clustering and adaptation effect that can be obtained via contrastive learning.
The row \emph{+ Style Transfer} shows a low performance for ROS when using this augmentation. By checking the predictions we observe a slight advantage in the recognition accuracy of the \emph{known} classes, but a significant drop in the \emph{unknown} accuracy which causes a decrease in the overall result. 
We also followed \cite{rakshit2020multi} to extend ROS with self-training. The corresponding row \emph{+ Self-Training} shows again a drop in performance:
this procedure tends to propagate the recognition errors due to the cross-entropy overconfidence.
Indeed self-training may induce a dangerous model drift, but recent literature has shown that its effectiveness and safe nature hold when the sample selection is performed with a self-pacing strategy based on the distribution of the unlabeled samples \cite{cascantebonilla2020curriculum}, exactly as in \OUR.

Finally, when applying all the strategies at once, the results are similar to those obtained with style transfer alone. This last technique clearly steered the whole method towards a low performance. 


\section{Extension to Closed-Set and Universal}
\OUR can be easily extended to the simpler multi-source Closed-Set domain adaptation setting (perfect overlap between sources and target classes) and to the more challenging multi-source Universal domain adaptation case (both sources and target have their own private categories). We consider the DomainNet dataset and run an evaluation on those two scenarios, following \cite{li2021dynamic} for Closed-Set and \cite{fu2020learning} for Universal. In the latter, sources and target share the first 150 classes in alphabetic order, the next 50 categories are sources private classes and the rest are target private classes. 
For Closed-Set we use as reference LtC-MSDA \cite{wang2020learning} and DRT \cite{li2021dynamic} which leverage respectively on a graph connecting domain prototypes, and on a dynamic transfer
that updates the model parameters on a per-sample  basis.
Table \ref{tab:Closed-Set} collects the results and show how \OUR gets promising performance with respect to several state-of-the-art methods in the two scenarios.

\section{Conclusions}
In this paper we introduced \OUR, a straightforward approach for multi-source Open-Set domain adaptation.
It exploits contrastive learning and the inherent properties of its hyperspherical feature space to avoid the limitations of the existing competing methods. \OUR includes a tailored data balancing to enforce cross-source alignment and introduces style transfer among the instance transformations for source-target adaptation, keeping away from the risk of negative transfer. Finally, a self-training strategy refines the model without the need for manually set thresholds. Through extensive experiments, we demonstrated state-of-the-art results on three  benchmarks and we delved into the details of the methods with several quantitative evaluations which shed light on its internal functioning. The application to the multi-source closed-set and universal scenario confirm the effectiveness of \OUR, identifying it a good starting approach towards life-long learning for real-world applications.

\noindent\textbf{Acknowledgements}. This work was partially
supported by the ERC project RoboExNovo. Computational resources were provided by IIT (HPC infrastructure). We also thank Biplab Banerjee for the discussions on MOSDANET \cite{rakshit2020multi}.

\section*{Appendix}
\appendix

\section{Qualitative Analysis}
\begin{figure*}[t!]
    \centering
\includegraphics[width=0.99\textwidth]{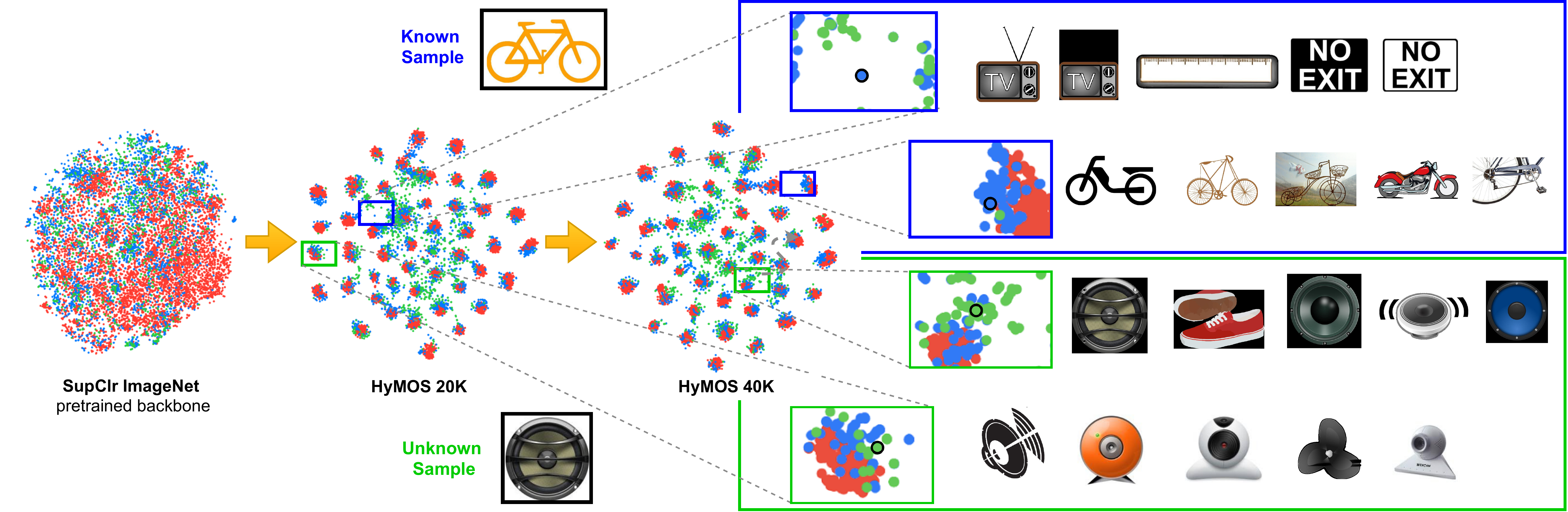}
    \caption{Qualitative analysis on the Ar,Pr,Rw $\rightarrow$ Cl case of the Office-Home dataset. The red dots represent the source domain, the blue dots are the known samples of the target domain, and the green dots are the unknown ones. 
    \OUR 20k: source balancing and style transfer already favor a good alignment of most of the known target classes with the respective source known cluster. \OUR 40k: self-training further move the target known samples towards the respective source clusters, while  the unknown samples remain in the regions among the clusters. The zooms show how the neighborhood of a known (bike) and unknown (speaker) target samples change during training. }
    \label{fig:tsne}
\end{figure*}

We visualize the distribution of source and target data in the feature space (output of the contrastive head) with the t-sne \cite{van2008visualizing} plots in Figure \ref{fig:tsne}. In particular, we focus on the Ar,Pr,Rw $\rightarrow$ Cl case of the Office-Home dataset: the red dots represent the source domain, the blue dots are the known samples of the target domain, and the green dots the unknown ones. We take three snapshots of the data on the hyperspherical embedding: at the beginning when the backbone network is inherited from SupClr \cite{NEURIPS2020_supclr} pre-trained on ImageNet, immediately before the first \emph{break-point} (\ie before the application of self-training), and at the end of the training process. 
By observing the intermediate plot we can state that source balancing and style transfer already favor a good alignment of most of the known (blue) target classes with the respective source known clusters (red). The  last plot indicates that self-training further improves the alignment while the unknown samples (green) remain in the regions among the clusters.

Randomly zooming on a known sample (the bike) and on an unknown sample (the speaker) we observe how their position change during training. The first moves from an isolated region where its top five neighbors show high class confusion, towards the correct bike class. The second starts from a neighborhood populated by several samples of classes webcam and fan and finally appears in a different region shared mostly by other instances of the class speaker.

\section{Further experiments}
\noindent\textbf{Complete results with additional metrics}
In Table \ref{tab:sota} we present the same results of the main paper including also additional metrics: the average class accuracy over known classes $OS^*$, the accuracy on the unknown class $UNK$ and the average accuracy over all classes $OS$ defined as $OS=\frac{|\mathcal{C}_s |}{|\mathcal{C}_s | + 1} \times {OS^*}+ \frac{1}{|\mathcal{C}_s | + 1} \times {UNK}$.

\begin{figure}[t!]
\centering
\hspace{-4mm}
\resizebox{0.27\textwidth}{!}{
\includegraphics[width=0.25\textwidth]{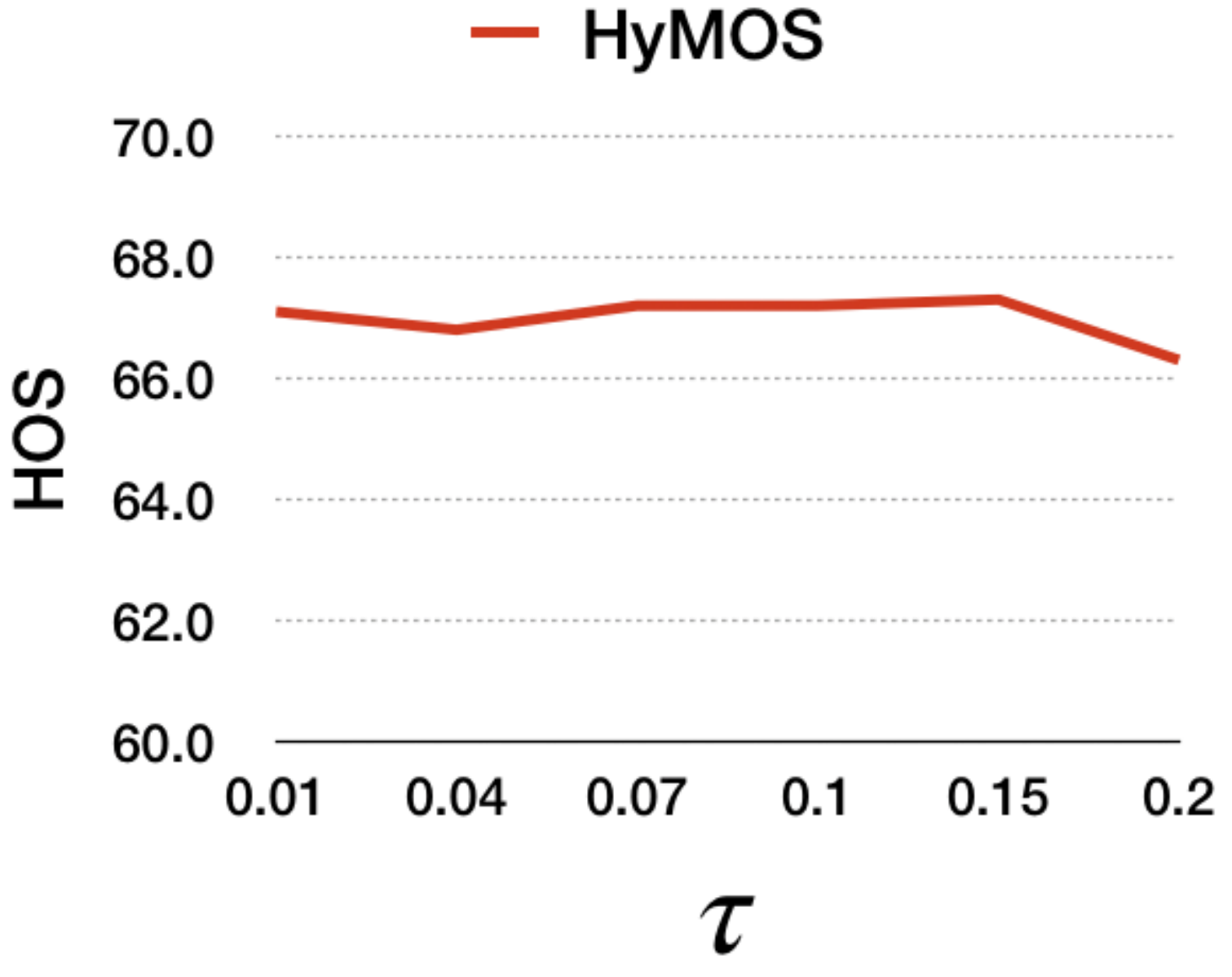}
}
\caption{Sensitivity analysis for the temperature value $\tau$ on Office-Home.}
\label{fig:ablationtau}\vspace{3mm}
\end{figure}

\noindent\textbf{Robustness to temperature variation}
The temperature $\tau$ in the contrastive loss (main paper Eq. (1)) is kept fixed to the default value  $0.07$ as suggested in \cite{tack2020csi}. We verified experimentally that the results are stable even when tuning $\tau$ and remain always higher than ROS (65.3) (see Figure \ref{fig:ablationtau}).

\begin{table*}[t]
    \centering
    
    \resizebox{\textwidth}{!}{
\begin{tabular}{cc cccc cccc cccc cccc| }
\hline

\multicolumn{18}{c|}{\textbf{Office31}}        \\
\hline

& & \multicolumn{4}{|c|}{D,A $\rightarrow$ W } & \multicolumn{4}{c|}{W,A $\rightarrow$ D } & \multicolumn{4}{c|}{W,D$\rightarrow$ A } &
\multicolumn{4}{c|}{Avg.}\\

 & & \multicolumn{1}{|c}{OS} & OS* & UNK & \multicolumn{1}{c|}{\textbf{HOS}} 
  &  OS & OS* & UNK & \multicolumn{1}{c|}{\textbf{HOS}} 
  &  OS & OS* & UNK & \multicolumn{1}{c|}{\textbf{HOS}} 
  &  OS & OS* & UNK & \multicolumn{1}{c|}{\textbf{HOS}}\\
 \hline
  
 \multicolumn{1}{c}{\multirow{4}{*}{Source Combine }} 
 
  & \multicolumn{1}{c|}{Inheritable  \cite{kundu2020towards}} & 	69.0 & 68.1 &	87.6 &	\multicolumn{1}{c|}{76.6}  & 74.7 &	74.1 &	85.6 &	\multicolumn{1}{c|}{79.5}  & 63.7 &	62.9 &	78.9 &	\multicolumn{1}{c|}{70.0}  & 69.1 &	68.4 &	84.0 &	75.4 \\
  
  &  \multicolumn{1}{c|}{ROS \cite{bucci2020effectiveness}}  & 82.2 &	82.3 &	81.5 &\multicolumn{1}{c|}{81.8}  &	95.3 & 96.5 &	68.7 &	\multicolumn{1}{c|}{80.1} &	53.8 & 52.2 &	84.9 &	\multicolumn{1}{c|}{64.7} &	77.1 &	77.0 &	78.4 &	75.5 \\
  
   &  \multicolumn{1}{c|}{CMU \cite{fu2020learning}} & 96.1 &	98.7 &	44.6 &	\multicolumn{1}{c|}{61.4}  & 96.2 &	98.7 &	47.3 &	\multicolumn{1}{c|}{64.0}  & 73.1 &	74.5 &	45.4 &	\multicolumn{1}{c|}{56.4}  & 88.5 &	90.6 &	45.8 &	60.6 \\
  
 &  \multicolumn{1}{c|}{DANCE \cite{saito2020dance}} & 95.9 &	99.5 &	23.9 &	\multicolumn{1}{c|}{38.5}  & 97.3 &	100.0 &	42.6 &	\multicolumn{1}{c|}{59.7}  & 78.0 &	79.6 &	45.6 &	\multicolumn{1}{c|}{58.0}  & 90.4 &	93.0 &	37.3 &	52.0  \\ 
 
 &  \multicolumn{1}{c|}{PGL \cite{pgl-luo20b-icml20}} & 94.1 & 97.4 & 27.8	 &	\multicolumn{1}{c|}{43.3}  & 92.2 &	95.6 &	23.5 &	\multicolumn{1}{c|}{37.7}  & 77.1 & 79.8 & 22.9	 &	\multicolumn{1}{c|}{35.6}  & 87.8 &	90.9 & 24.7 & 38.9\\

  \hline

 \multicolumn{1}{c}{\multirow{2}{*}{Multi-Source }} & \multicolumn{1}{c|}{MOSDANET \cite{rakshit2020multi}}  & 97.7 &	99.4 &	43.5 &	\multicolumn{1}{c|}{60.5}  & 97.0 &	99.0 &	55.9 &	\multicolumn{1}{c|}{71.5}  & 80.9 &	81.5 &	67.6 &	\multicolumn{1}{c|}{\textbf{73.9}}  & 91.9 & 93.3 &	55.7 & 68.6\\
 
 & \multicolumn{1}{c|}{\textbf{\OUR}}  & 96.1 &	96.6 &	84.6 &	\multicolumn{1}{c|}{\textbf{90.2}}  & 96.7 & 97.3 &	83.6 &	\multicolumn{1}{c|}{\textbf{89.9}}  & 49.6 & 48.0 &	83.1 &	\multicolumn{1}{c|}{60.8}  & 80.8 &	80.6 &	83.8 &	\textbf{80.3}\\

\hline
\end{tabular}
\begin{tabular}{cccc cccc cccc  }
\hline

\multicolumn{12}{|c}{\textbf{DomainNet}}        \\
\hline

 \multicolumn{4}{|c|}{I,P $\rightarrow$ S } & \multicolumn{4}{c|}{I,P $\rightarrow$ C } & \multicolumn{4}{c}{Avg.}\\

 \multicolumn{1}{|c}{OS} & OS* & UNK & \multicolumn{1}{c|}{\textbf{HOS}} 
  & OS & OS* & UNK & \multicolumn{1}{c|}{\textbf{HOS}} 
  & OS & OS* & UNK & \multicolumn{1}{c}{\textbf{HOS}}\\
 \hline
  	\multicolumn{1}{|c}{24.9} &	24.5 & 60.3 &	\multicolumn{1}{c|}{34.8}  & 33.5 &	33.1 &	65.6 &	\multicolumn{1}{c|}{44.0}  & 29.2 &	28.8 &	62.9 &	39.4 \\
  
   	\multicolumn{1}{|c}{31.7} & 31.3 &	77.5 &	\multicolumn{1}{c|}{44.5} &	41.0 & 40.7	 & 73.6 &	\multicolumn{1}{c|}{52.4}  &	36.4 & 36.0 &	75.5 &	48.5 \\
  
 \multicolumn{1}{|c}{48.0} & 48.3 &	26.3 &	\multicolumn{1}{c|}{38.1}  &	49.6 & 49.8 &	27.6 &	\multicolumn{1}{c|}{35.5} &	48.8 & 49.1 &	27.0 &	36.8\\  
  
\multicolumn{1}{|c}{45.6} & 45.8 &	22.3 &	\multicolumn{1}{c|}{30.0}  & 54.4 &	54.7 &	28.7 &	\multicolumn{1}{c|}{37.6}  & 50.0 &	50.3 &	25.5 &	33.8 \\ 

\multicolumn{1}{|c}{54.9} & 55.3 & 	 11.1 &	\multicolumn{1}{c|}{18.5}  & 59.6 &	60.1 & 11.6	 &	\multicolumn{1}{c|}{19.4}  & 57.3 &	57.7 & 11.4	 & 19.0 \\

  \hline

	\multicolumn{1}{|c}{30.2} & 29.9 &	60.2 &	\multicolumn{1}{c|}{40.0}  & 31.8 &	31.6 &	51.8 &	\multicolumn{1}{c|}{39.3}  & 31.0 &	30.8 &	56.0 &	39.6\\
 
 \multicolumn{1}{|c}{43.6} & 43.2 &	86.0	 & \multicolumn{1}{c|}{\textbf{57.5}}  & 47.8 &	47.4 &	85.5 &	\multicolumn{1}{c|}{\textbf{61.0}}  & 45.7 &	45.3 &	85.8 &	\textbf{59.3}\\

\hline
 
\end{tabular}
}

 \resizebox{\textwidth}{!}{

\begin{tabular}{cc cccc cccc cccc cccc cccc}
\hline

\multicolumn{22}{c}{\textbf{Office-Home}}        \\
\hline

& & \multicolumn{4}{|c|}{Ar,Pr,Cl $\rightarrow$ Rw } & \multicolumn{4}{c|}{Ar,Pr,Rw $\rightarrow$ Cl } & \multicolumn{4}{c|}{Cl,Pr,Rw $\rightarrow$ Ar } & \multicolumn{4}{c|}{Cl,Ar,Rw $\rightarrow$ Pr } & 
\multicolumn{4}{c}{Avg.}\\

 &  &  \multicolumn{1}{|c}{OS} & OS* & UNK & \multicolumn{1}{c|}{\textbf{HOS}} 
   & OS  & OS* & UNK & \multicolumn{1}{c|}{\textbf{HOS}} 
   & OS  & OS* & UNK & \multicolumn{1}{c|}{\textbf{HOS}} 
   & OS  & OS* & UNK & \multicolumn{1}{c|}{\textbf{HOS}}
   & OS  & OS* & UNK & \multicolumn{1}{c}{\textbf{HOS}} \\
 \hline
 \multicolumn{1}{c}{\multirow{4}{*}{Source Combine }} 
 
  & \multicolumn{1}{c|}{Inheritable \cite{kundu2020towards}} & 58.6 & 	58.4 &	68.9 &	\multicolumn{1}{c|}{63.2}  & 44.3 &	43.7 &	66.5 &	\multicolumn{1}{c|}{52.6}  & 36.4 &	35.5 &	77.6 &	\multicolumn{1}{c|}{48.7}  & 58.6 &	58.5 &	63.3 &	\multicolumn{1}{c|}{60.7}  & 49.5 &	49.1 &	69.1 &	56.3\\
  
  &  \multicolumn{1}{c|}{ROS \cite{bucci2020effectiveness}} & 	69.9 & 69.8 &	76.9 &	\multicolumn{1}{c|}{\textbf{73.0}} & 57.1 &	57.1 &	57.6 &	\multicolumn{1}{c|}{57.3}  & 57.5 &	57.2 &	66.7 &	\multicolumn{1}{c|}{61.6}  & 70.3 &	70.3 &	68.0 &	\multicolumn{1}{c|}{69.1} &	63.7 & 63.6 &	67.3 &	65.3\\
  
  &  \multicolumn{1}{c|}{CMU \cite{fu2020learning}} & 62.9 &	62.5 &	81.5 &		\multicolumn{1}{c|}{70.8} &	35.8 & 34.6 &	89.9 &		\multicolumn{1}{c|}{50.0}  & 44.6 &	43.7 &	87.0 &		\multicolumn{1}{c|}{58.1}  & 60.6 &	60.1 &	81.7 &		\multicolumn{1}{c|}{69.3}  & 51.0 &	50.2 &	85.0 &	62.1 \\ 
  
 &  \multicolumn{1}{c|}{DANCE \cite{saito2020dance}} & 83.9  &	85.6 &	4.5 &	\multicolumn{1}{c|}{12.4} & 66.8 &	68.0 &	9.2 &	\multicolumn{1}{c|}{16.1}  & 72.7 &	74.1 &	10.7 &	\multicolumn{1}{c|}{18.6}  & 85.1 &	86.7 &	13.4 &	\multicolumn{1}{c|}{22.9}  & 77.1 &	78.6 &	9.4	 & 17.5 \\
 
  &  \multicolumn{1}{c|}{PGL \cite{pgl-luo20b-icml20}} & 83.4  &	84.6 &	26.2 &	\multicolumn{1}{c|}{40.0} & 62.0 & 63.0 & 21.0 &	\multicolumn{1}{c|}{31.5}  & 69.5  & 70.6 &	20.5 &	\multicolumn{1}{c|}{31.8}  & 82.6  & 83.8 &	28.2 &	\multicolumn{1}{c|}{42.2}  & 74.4  & 75.5	 &	24.0	& 36.4 \\

 \hline
 
 \multicolumn{1}{c}{\multirow{2}{*}{Multi-Source }} & \multicolumn{1}{c|}{MOSDANET \cite{rakshit2020multi}} &	78.4 & 79.4 &	55.0 &	\multicolumn{1}{c|}{65.0} & 67.5 &	68.1 &	40.9 &	\multicolumn{1}{c|}{51.1}  & 61.0 &	61.3 &	48.7 &	\multicolumn{1}{c|}{54.3}  & 81.1 &	82.2 &	55.0 &	\multicolumn{1}{c|}{65.9}  & 72.0 &	72.8 &	49.9 &	59.1\\
 
 & \multicolumn{1}{c|}{\textbf{\OUR}}  & 69.5 &	69.4 &	72.7 &		\multicolumn{1}{c|}{71.0}  & 52.5 &	51.7 &	86.0 &		\multicolumn{1}{c|}{\textbf{64.6}}  & 50.1 &	49.4 &	84.1 &		\multicolumn{1}{c|}{\textbf{62.2}}  & 71.5 &	71.5 &	70.6 &		\multicolumn{1}{c|}{\textbf{71.1}} & 60.9 &	60.5 &	78.4 &	\textbf{67.2} \\

\hline
\end{tabular}
}   

\vspace{3mm}\caption{Accuracy (\%) averaged over three runs for each method on the Office31, DomainNet and Office-Home datasets.}
\label{tab:sota}
\end{table*}

\section{Implementation Details}

We implemented HyMOS with an architecture composed of the ResNet-50 \cite{resnet} backbone that corresponds to the \emph{encoder} and two fully connected layers of dimension 2048 and 128 which define the \emph{contrastive head}. The overall network is trained by minimizing the contrastive loss (see the main paper, Eq. (1)), setting $\tau=0.07$ as in \cite{tack2020csi}. Our distance-based classifier lives in the hyperspherical space produced by the model, whose dimension is not constrained by the number of classes. As a consequence, the architecture remains exactly the same for all our experiments.

We initialize the backbone network with the ImageNet pre-trained SupClr model \cite{NEURIPS2020_supclr} and train \OUR for 40k iterations with a balanced data mini-batch which contains one sample for each class of every source domain. 
The learning rate grows from 0 to 0.05 (at iteration 2500) with a linear warm-up schedule, to then decrease back to 0 at the end of training (iteration 40k) through a cosine annealing schedule. We use LARS optimizer \cite{LARS} with momentum 0.9 and weight decay $10^{-6}$. For the first 20k iterations we train only on source data, using target data exclusively for the style transfer based data augmentation for the supervised contrastive learning objective. We then perform an eval step that we call self-training \emph{break-point} in order to start including confident known target samples in the learning objective. We perform \emph{break-point} eval steps every 5K iterations till the end of the training.

\setlength{\textfloatsep}{0pt}
\begin{algorithm}[tb]
\caption{\OUR evaluation procedure}
\label{alg:eval}
\begin{algorithmic}
\Require $\sT$; trained $Enc$ and $Proj$
\Ensure Predictions on $\sT$

\Procedure{finalEval()}{}
\State $\alpha \leftarrow $ (main paper Eq. (4))
\For{\textbf{each} $\bx_t$ \textbf{in} $\sT$}
\State $\bz^t = Proj(Enc(\bx^t)$)
\State $ h_{y^s} \leftarrow \texttt{nearest prototype to } \bz^t  $
\If{$ d_{h_{y^s}}(z^t)  < \alpha$ }
\State $ \hat{y}^t = y^s $
\Else
\State $ \hat{y}^t = \texttt{unknown} $
\EndIf
\EndFor
\EndProcedure

\Procedure{main()}{}
\State $finalEval()$
\EndProcedure

\end{algorithmic}
\end{algorithm}

For style transfer data augmentation we use the standard VGG19-based AdaIN model with default hyperparameters \cite{huang2017adain}, trained with content data from the available source domains and target samples as style data.

For what concerns the instance transformations, we applied the same data augmentations originally proposed for SimClr \cite{simclr2020}, extending them with style transfer. Specifically, we used random resized crop with scale in $\{0.08, 1\}$ and random horizontal flip. The style transfer is applied with probability $p=0.5$ on the source images, while the remaining not-stylized images are transformed via color jittering with probability $p=0.8$ and grayscale with probability $p=0.2$.

The final evaluation procedure of \OUR is summarized in Algorithm \ref{alg:eval}.

{\small
\bibliographystyle{ieee_fullname}
\bibliography{egbib}
}

\end{document}